# Comparison of Algorithms for Compressed Sensing of Magnetic Resonance Images

Badnjar Jelena
Student, Faculty of Electrical Engineering, University of Montenegro
Podgorica, Montenegro
jelena.badnjar@t-com.me

*Abstract*— **Magnetic resonance imaging (MRI) is an essential medical tool with inherently slow data acquisition process. Slow acquisition process requires patient to be long time exposed to scanning apparatus. In recent years significant efforts are made towards the applying Compressive Sensing technique to the acquisition process of MRI and biomedical images. Compressive Sensing is an emerging theory in signal processing. It aims to reduce the amount of acquired data required for successful signal reconstruction. Reducing the amount of acquired image coefficients leads to lower acquisition time, i.e. time of exposition to the MRI apparatus. Using optimization algorithms, satisfactory image quality can be obtained from the small set of acquired samples. A number of optimization algorithms for the reconstruction of the biomedical images is proposed in the literature. In this paper, three commonly used optimization algorithms are compared and results are presented on the several MRI images.**

## I. Introduction

MRI, as medical imaging tool, usually takes a long time to collect information related to the patient body condition which adversely affects the physical condition of the patient. Also, having large number of samples complicate the real-time analysis and requires large storage capacities. Recent research efforts in signal processing show the possibility to recover the signal using significantly smaller amount of collected information. According to this, signals can be acquired with a sampling rate far smaller than the Nyquist rate, without any loss of quality. This approach is applicable both, for 1D and 2D signals, and is called Compressive Sensing (CS).

In general, CS approach requires a priori defined conditions to be satisfied in order to be applicable to certain signal. Important property signal should satisfy is to have representation in its own domain or in the transformation domain (Fourier domain, discrete cosine transform domain, wavelet domain, etc.), that has small number of non-zero samples. Such signal representation is known as a sparse representation. There should also exist a domain where signal is dense and in which CS acquisition procedure is performed.

Sparse signal may be dense in the time domain like e.g. sinusoid, and after transforming it into frequency domain, signal is fully represented with one non-zero sample that corresponds to the sinusoid frequency. Most of magnetic resonance images can be sparsified in e.g. wavelet domain while others have a sparse temporal Fourier transform [1]. The second important fact that opens the door for CS in MRI is energy. Energy is concentraced close to the center of 2D (or 3D) Fourier transform of the MR image measured [1]. This space, called k-space, is in frequency domain and represents raw image data before reconstruction. Its complex values are sampled during an MR measurement and stored in matrix, during data acquisition. Designing a CS scheme for MRI can be viewed as selecting a subset of the frequency domain [1]-[4]. Designs should have variable-density sampling with denser sampling near the center of k-space, matching the energy distribution [4].

Signal in CS should be acquired in incoherent way, in order to be recoverable using small set of collected samples. Random sampling satisfies incoherence property and this is the most common way of acquiring signal in CS approach. CS allows reconstruction of the signal having small number of known signal samples, by using optimization algorithms. Nowadays, CS have been applied in various applications [5]-[11]: radars, communications, multimedia, image restoration, etc. Optimization algorithms are based on different norms minimization, depending on the type of the signal and application. The most commonly used is $l_1$-norm minimization. Having in mind the nature of 2D signals, the reconstruction is usually done by using total variation method [12]. Total variation is minimization of the image gradient and it is commonly used method for the 2D signal reconstruction.

The paper is structured as follows: Second part contains theoretical background of the CS approach. Three commonly used algorithms for the MRI images reconstruction are presented in the Section III. Section IV shows performances of the algorithms and comparison in terms of execution time and peak signal to noise ratio (PSNR). The results are summarized in the Conclusion part.

## II. Theoretical background

Unlike the traditional sampling based on the Shannon Nyquist sampling theorem, CS proposes the acquisition of the signal with significantly smaller number of samples compared to the traditional sampling. It allows signal reconstruction from this small number of acquired samples, by using optimization algorithms [13]-[16]. Let us describe basic concepts of the CS on the discrete signal $x$ of length $N$. If the $N$-dimensional signal $x$, can be represented using $K$ samples in the certain domain, it is said that the signal is *K*-sparse in this domain. The case of interest is when $K << N$. Let define one dimensional signal x $\in$ $R^N$ in the form of the basis vectors $S_i$ and transform matrix $\Psi$:

$$x = \sum_{i=1}^{N} \Psi_i S_i \qquad (1)$$

Suppose *x* is *K* sparse in basis $\Psi$, so x=S$\Psi$, with $\|S\|_0$ = K. Term $\|.\|_0$ represents $l_0$-norm and returns the number of nonzero elements. Sampling can be done using the sampling (measurement) matrix $\Phi$, of dimension $M \times N$. The result of the sampling with the matrix $\Phi$ is the measurement vector *y*:

$$y = \Phi x = \Phi \Psi S = \theta S \qquad (2)$$

Matrix $\theta$ is called the CS matrix. It can be viewed as the sub-matrix of the transform domain matrix $\Psi$, as it is formed by choosing certain number of randomly permuted row/columns of the matrix $\Psi$. Choosing rows/columns randomly, it is assured that the incoherence property is satisfied. Equation (2) represents system with *M* equations and *N* unknowns, where *M<N*. Therefore, system is undetermined and has infinite number of solutions. This problem can be solved using optimization techniques. Optimization problem is defined as:

$$\hat{x} = \min \|x\|_{l_1} \qquad (3)$$

with condition y=$\Phi$x, where $\hat{x}$ is solution of minimization problem and $\ell_1$ norm of the vector x is defined as:

$$\|x\|_{l_1} = \sum_{i=1}^{N} |x_i| \qquad (4)$$

Having in mind that the images are not sparse in frequency and space domain, the optimization algorithm is done by minimizing the total variation (TV) of the 2D signal. If we consider that *S* is 2D signal, than the TV minimization problem is defined as:

$$\min_{S} TV(S) \text{ subject to } y = \theta S, \qquad (5)$$

where the TV of the pixel at the *ij* position is described by using the following relation:

$$TV(S) = \sum_{i,j} \|S_{i+1,j} - S_{i,j}, S_{i,j+1} - S_{i,j}\|_2 \qquad (6)$$

I.e. TV represents sum of the gradient approximation for the each pixel position.

### III. CS Reconstruction Algorithms

CS imaging relies on the ability to efficiently solve a convex minimization problem. As previously mentioned, minimization of the TV is used rather than the $\ell_1$ norm minimization. In the sequel, three commonly used optimization algorithms in biomedical applications are reviewed and their performances are compared.

*A. TwIST*

Two-step Iterative Shrinkage/Tresholding algorithm, has the task to manage with convex optimization problems. TwIST represents modification of the Iterative Shrinkage/Tresholding algorithm, and shows to be faster compared to the IST, for the wavelet-based and TV-based image restoration problems.

TwIST restores MR images using two-step algorithm in which every estimate depends on the two previous estimates rather than only previous one. TwIST converges to a minimizer of the objective function, which is convex and defined as:

$$f(x) = \frac{1}{2} \|y - Kx\|^2 + \lambda \Phi(x), \qquad (7)$$

where K is the linear operator that maps Hilbert spaces $\chi \to \Upsilon$, $\lambda$ is the regularization parameter, and $\phi$ is the convex regularizer: $\chi \to R$ and it can be TV or $l_1$ regularization. According to the linear system, given by $Ax = K^T y$, the two-step iteration becomes:

$$x_{t+1} = (1-\alpha)x_{t-1} + (\alpha-\beta)x_t + \beta C^{-1}(x + K^T(y - Kx_t)) \qquad (8)$$

This resulted from taking C=I+$D_t$, where D is an orthogonal matrix which depends on x and $\Phi$ and R=I-$K^T$K, we get A=C-R, remarking I represents K orthogonal for pure denoising problem. As we indicated, when x represents wavelet coefficients of the image, $\Phi(x)=\|x\|$ is regularizer.

*B. RecPF*

The RecPF (Reconstruction from Partial Fourier algorithm) algorithm is proposed by Yang, Zhang, and Yin [17], and was designed for the reconstruction of the MRI images. RecPF solves CS reconstruction problems using $l_1$-norm or TV (or both) and $\Phi$ as a subsample of the Fourier transformation [17]. RecPF includes shrinkage and Fast Fourier transform or discrete cosine transform, at each iteration. The algorithm solves the following problem:

$$\min_{x \in R^N} \|x\|_{TV} + \frac{\lambda}{2} \|Kx - y\|_{l_1}^2, \qquad (9)$$

where parameters in the equations (7), (3) and (5) correspond to the parameters in the (9). Due to the algorithm high perofomances in terms of reconstruction speed and quality, RecPF can be used in the future in sparsity-based, fast MRI reconstruction.

*C. SALSA*

Split Augmented Lagrangian shrinkage algorithm involve reconstruction from partial Fourier transform. It includes reconstruction and deblurring/deconvolution from compressive observations using either total-variation or wavelet-based regularization. SALSA is based on the technique known as variable splitting to obtain an augmented Lagrangian method ([3],[18]). As noted in Section II, an unknown image x can be presented as a linear combination of the elements of some frame, i.e., x=$\Psi$S, where the columns of the matrix $\Psi$ are the elements of a wavelet frame. Then, the coefficients of this representation are estimated from the noisy image, under the $\ell_1$ norm. This regularization process is known as wavelet-based, assigned to image deblurring. An alternative formulation applies a regularizer directly to the unknown image, which analyzes the image itself, rather than the coefficients of a representation thereof. The most often used regularizer in this case is the total variation, assigned to image restoration.

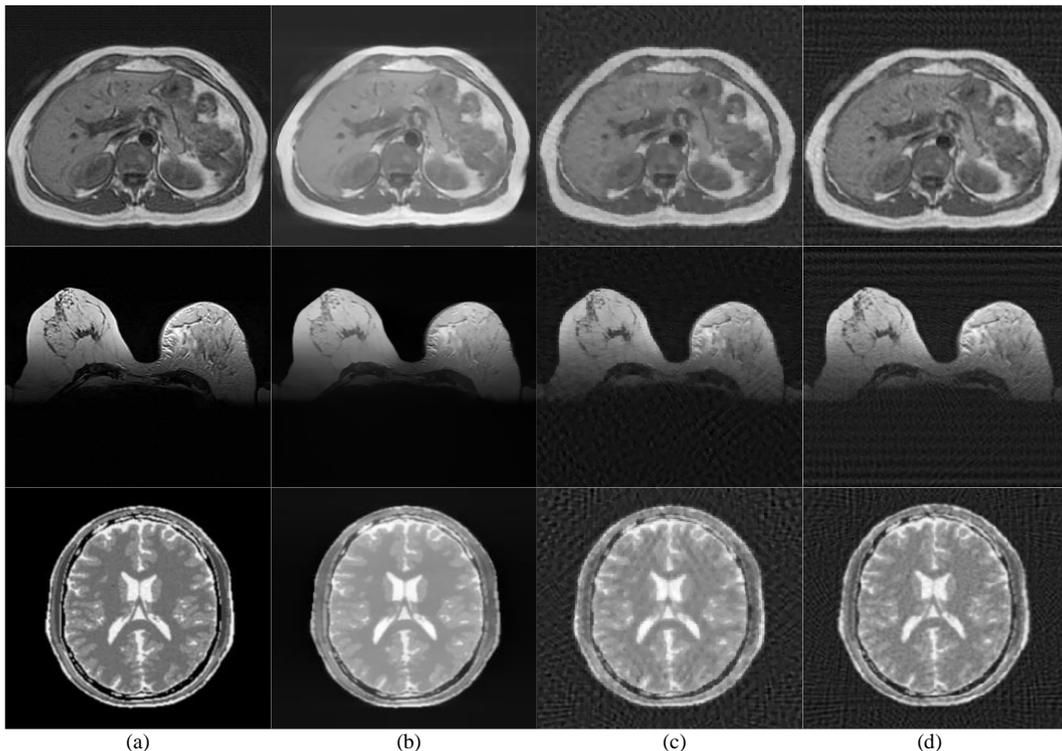

(a)        (b)        (c)        (d)

Figure 1. Magnetic resonance reconstruction result of liver image (top row), breast image (middle row) and brain image (bottom row). Original images in first column (a) and estimated images using RecPF (b), SALSA (c) and TwIST (d), measuring 25% of coefficients.

*D. Comparison of reconstruction algorithms*

First we compare RecPF with TwIST. RecPF solves the following problem: $\min_x \lambda^{-1}\Phi_{reg}(x) + \frac{1}{2}\|Kx - y\|^2$ where $\Phi_{reg}$ can be both TV or $\ell_1$ regularization and K is linear operator. Signal y contains enough information for successful reconstruction. Now, iteration framework of TwIST becomes: $x_{t+1}=(1-\alpha)x_{t-1} + (\alpha-\beta)x_t+\beta\Psi_\lambda(\varepsilon_t)$, where $\alpha,\beta>0$ are parameters, $\varepsilon_t=x_t+K^T(b-Kx_t)$ with $b=Ax$ and

$$\Psi_\lambda(\varepsilon_t) = \arg\min_x \lambda^{-1}\Phi_{reg}(x) + \frac{1}{2}\|x - \varepsilon_t\|^2 \quad (10)$$

Problem (10) is solved iteratively by Chambolle's algorithm in TwIST and SALSA, while RecPF solves TV problem at an approximate cost of 2FFTs [3]. In terms of the iteration and CPU time consumed, RecPF is relatively stable as λ varies while TwIST appears to be sensitive [17]. RecPF with the $\ell_1$ term dropped has a per-iteration cost of 2 FFTs (including 1 inverse FFT) which is much lower than solving by Chamolle's algorithm. This is one of the main reasons that RecPF runs faster. TwIST cannot solve minimization problem with both the TV and $\ell_1$ regularization terms. RecPF minimizes a potential (objective) function with one or both TV and $\ell_1$ regularization terms. As in the case of some of the image deconvolution problems with orthogonal wavelets, it may be possible to get a solution using TwIST, quicker than SALSA [3]. Slowness in TwIST in relation to SALSA can be caused by the use of a small value of the regularization parameter until SALSA keeping parameter fixed, then updating it and repeating these two steps until some convergence criterion is satisfied [3]. Value of regularizer parameter in RecPF can be difficult to tune and this algorithm is limited to Fourier sampling.

IV. EXPERIMENTAL RESULTS

In this section, the experimental results using three common algorithms for MRI image reconstruction are presented. The small database of the MRI images is formed (three real MRI images are used for testing). Images were reconstructed by using RecPF, SALSA and TwIST algorithm. All of the algorithms use TV minimization to solve optimization problem and take measurements from frequency domain. For all experiments, we used the same sampling pattern with radial lines, choosing 25% of the total number of image coefficients. Results are presented in Fig.1.

We have tested reconstruction on brain, liver and breast images. The results are summarized in Table I. Relative reconstruction error and peak signal to noise ratio (PSNR) are calculated for each image. Observing the reconstructed images, it can be seen that TwIST algorithm gives the best results, considering image details. PSNR varies slightly for between the used algoritms, when observing each individual image. However, SALSA algorithm gives the highest PSNR values in all considered cases. We can also notice that there are differences between the algorithm execution times: for the same input, RecPF can be roughly 22 times faster than SALSA and roughly 23 times faster than TwIST. In terms of iteration numbers, TwIST reconstruction algorithm measuring 25% of coefficients, needs 15 iterations for each MR image. According to this, RecPF needs 30 iterations to reconstruct images until SALSA needs 93 iterations for brain image, 89 for liver image and 98 iterations for breast image. It is obvious that the each algorithm is good representation for CS as reconstruction and denoising tecnique, where TwIST needs the lowest number of iterations with good PSNR results.

TABLE I

Reconstruction Relative Errors and Peak-Signal-to-Noise ratios

| Image<br>Alg. | | Breast (25%) | Brain (25%) | Liver (25%) |
|---|---|---|---|---|
| RecPF | *Rel. error (%)* | 15.99 | 17.00 | 6.86 |
| | *PSNR(dB)* | 27.98 | 24.41 | 29.02 |
| | *CPU time(sec)* | 0.445 | 0.441 | 0.447 |
| SALSA | *Rel. error (%)* | 12.51 | 15.16 | 4.03 |
| | *PSNR(dB)* | 30.11 | 25.36 | 32.62 |
| | *CPU time(sec)* | 10.00 | 9.29 | 8.06 |
| TwIST | *Rel. error (%)* | 14.19 | 15.89 | 11.78 |
| | *PSNR(dB)* | 29.02 | 24.92 | 29.23 |
| | *CPU time(sec)* | 10.51 | 10.69 | 10.32 |

CONCLUSION

In this paper, we have proposed a short review of recent CS imaging applications for biomedical applications as magnetic resonance imaging. The aim is to reach the fastest and most reliable reconstruction from as little samples as possible. The ultimate goal is to design MR scanner with integrated CS, which will reduce scan time, with benefits for patients, and energy consumption.

We also compared RecPF, TwIST and SALSA optimization algorithms for MR image reconstruction where experiments have illustrated the reconstruction capabilities of the CS. As can be seen from Table I, SALSA has the best reconstruction quality measure(PSNR) until TwIST requires the smallest iteration number with PSNR values slightly less than SALSA. RecPF, compared to other two algorithms, for the shortest consumed CPU time, reconstruct images with similar PSNR values as obtained in TwIST algorithm.

V. ACKNOWLEDGMENT

The authors are thankful to Professors and assistants within the Laboratory for Multimedia Signals and Systems, at the University of Montenegro, for providing the ideas, codes, literature and results developed for the project CS-ICT (funded by the Montenegrin Ministry of Science).